\pdfoutput=1

\documentclass[11pt]{article}

\usepackage[]{acl}

\usepackage{times}
\usepackage{latexsym}

\usepackage[T1]{fontenc}

\usepackage[utf8]{inputenc}

\usepackage{microtype}

\usepackage{tabularx}
\usepackage{multirow}
\usepackage{bm}
\usepackage{amssymb}
\usepackage{graphicx} 
\usepackage{comment}
\usepackage{booktabs}
\usepackage{chngpage}
\usepackage{xcolor}
\usepackage {tablefootnote}

\title{Argument Mining as a Text-to-Text Generation Task}

\author{Masayuki Kawarada${}^{1}$, Tsutomu Hirao${}^{2}$, Wataru Uchida${}^{1}$ \and Masaaki Nagata${}^{2}$\\
  ${}^{1}$Service Innovation Department, NTT DOCOMO, INC., Japan \\
  ${}^{2}$NTT Communication Science Laboratories, NTT Corporation, Japan \\
  \texttt{\{masayuki.kawarada.vw, uchidaw\}@nttdocomo.com} \\
  \texttt{\{tsutomu.hirao, masaaki.nagata\}@ntt.com} 
}

\begin{document}
\maketitle
\begin{abstract}
Argument Mining~(AM) aims to uncover the argumentative structures within a text. Previous methods require several subtasks, such as span identification, component classification, and relation classification. 
Consequently, these methods need rule-based postprocessing to derive argumentative structures from the output of each subtask. 
This approach adds to the complexity of the model and expands the search space of the hyperparameters. 
To address this difficulty, we propose a simple yet strong method based on a text-to-text generation approach using a pretrained encoder-decoder language model. 
Our method simultaneously generates argumentatively annotated text for spans, components, and relations, eliminating the need for task-specific postprocessing and hyperparameter tuning. 
Furthermore, because it is a straightforward text-to-text generation method, we can easily adapt our approach to various types of argumentative structures.
Experimental results demonstrate the effectiveness of our method, as it achieves state-of-the-art performance on three different types of benchmark datasets: the Argument-annotated Essays Corpus~(AAEC), AbstRCT, and the Cornell eRulemaking Corpus~(CDCP).\footnote{Our source code can be found at \url{https://github.com/masayouk/am-as-text2text}.}

\end{abstract}

\begin{figure*}[t]
\centering
  \includegraphics[width=\linewidth]{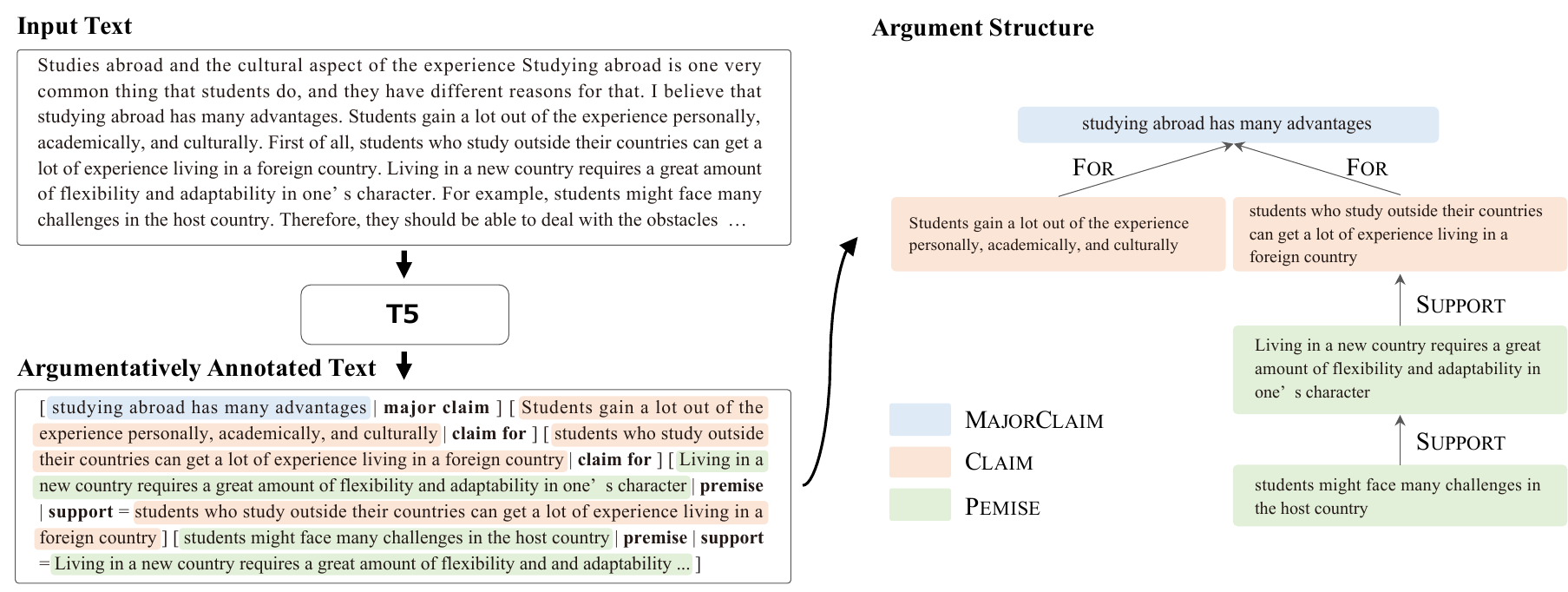}
  \caption{Overview of our methods. For our methodology, we input text into a pretrained encoder-decoder, such as T5 and FLAN T5. This process generates an argumentatively annotated text with spans, components, and relations. We then postprocess the output text to extract the argumentative structure.}
  \label{fig:method_overview}
\end{figure*}

\section{Introduction}
Argument Mining (AM) is a form of discourse analysis that seeks to identify the structure of an argument within a text~\cite{lawrence-reed-2019-argument}.
This structure is typically represented through a dependency tree or a directed acyclic graph, as shown on the right-hand side of Figure~\ref{fig:method_overview}.
In the dependency tree, nodes correspond to text spans that contain arguments, which are then classified into specific argument types. The edges between the nodes 
represent the relations between the arguments.

Annotated corpora have been constructed for argument mining to reveal the argumentative structure across various fields, such as student essays~\cite{stab-gurevych-2014-annotating, stab-gurevych-2017-parsing}, biomedical research~\cite{mayer:hal-02879293}, and more. 
These corpora serve as the standard benchmark datasets, allowing for the performance evaluation of argument mining systems.
Given its practical applications in downstream tasks, such as text summarization~\cite{fabbri-etal-2021-convosumm, elaraby-litman-2022-arglegalsumm} and automatic essay scoring~\cite{nguyen_essay}, argument mining has recently gained significant attention in discourse analysis.

Neural models enhanced the performance of argument mining, along with other natural language processing tasks.
Early models employed a pipeline approach involving three subtasks: identifying the argumentative text span, determining the argument type, and establishing the relation between the two arguments~\cite{stab-gurevych-2017-parsing, niculae-etal-2017-argument}. 
However, recent models treat argument mining as dependency parsing and perform it in an end-to-end manner~\cite{ye-teufel-2021-end,morio-etal-2022-end}. 
These models are complex as they require separate mechanisms for each of the three tasks to be incorporated into the model. 
Therefore, postprocessing is necessary to build valid dependency trees. 
Furthermore, hyperparameter tuning poses difficulties in implementing these models.

To tackle these difficulties, we exploited a simple text-to-text generation model with the Translation between Augmented Natural Languages~(TANL)~\cite{tanl}, which has achieved state-of-the-art performance on sentence-level structured prediction tasks such as relation extraction, named entity recognition, and semantic role labeling. 
Implementing TANL into AM, we offer significant advantages: (1) a simple architecture that eliminates complex postprocessing and hyperparameter tuning, (2) the ability to adapt to various annotations based on the dataset, and (3) the potential to use recent large language models.

Experimental results from three benchmark datasets, Argument-annotated Essays Corpus~(AAEC)~\cite{stab-gurevych-2017-parsing}, AbstRCT~\cite{mayer:hal-02879293} and Cornell eRulemaking Corpus~(CDCP)~\cite{park-cardie-2018-corpus}, demonstrate that our method achieved the state-of-the-art scores on both Component-F1 and Relation-F1 when using FLAN T5~\cite{FLANt5}:XXL (11B).
Furthermore, by preventing the model from generating irrelevant text spans, which cannot be arguments, we successfully reduced the computational time for inference in AbstRCT by 30\% without compromising performance.

\section{Related Work}
\subsection{Argument Mining}
AM involves three critical subtasks: identifying the arguments within a text, determining their argument type, and establishing the relations between these arguments. 
These steps are crucially required in revealing the argumentative structure of a text.
Earlier methods used a pipeline architecture, where argumentative span identification was performed first, followed by component classification\footnote{An argumentative text span assigned with a label, such as Claim or Premise, is called a component.} and relation classifications~\cite{persing-ng-2016-end,eger-etal-2017-neural, kuribayashi-etal-2019-empirical, morio-etal-2020-towards}.
However, such an approach can result in the accumulation of errors from previous subtasks.

To improve the pipeline-based approach, recent studies employ an end-to-end method~\cite{morio-etal-2022-end, bao-etal-2022-generative, ye-teufel-2021-end,eger-etal-2017-neural}. 
\citet{ye-teufel-2021-end} and \citet{morio-etal-2022-end} used a network architecture based on a biaffine parser, which achieved state-of-the-art performance on paragraph and essay level evaluation.
This approach treats the argumentative structure as a dependency tree and uses a dependency parsing algorithm to parse them.
Despite being end-to-end models, they often require hand-crafted rules~\cite{eger-etal-2017-neural, ye-teufel-2021-end} or an optimum branching algorithm~\cite{morio-etal-2022-end} to form dependency trees from the outputs of three layers corresponding to subtasks. Additionally, these models present challenges in tuning hyperparameters such as the learning rate, given the embedding of three subtasks within a network.

By contrast, \citet{bao-etal-2022-generative} use an encoder-decoder model to perform AM as a generation task. 
They employ a constraint pointer-mechanism~(CPM) for BART~\cite{lewis-etal-2020-bart} to predict the index of words in the input text.
However, our work differs from theirs as we focus on the text-to-text generation task.
This enables us to maximize the use of the decoder without making any modifications to the pretrained language model.

\subsection{Information Extraction as a Generation Task}
The recent development of pretrained language models~\cite{2020t5, lewis-etal-2020-bart} has led researchers to tackle information extraction tasks such as relation extraction~\cite{huguet-cabot-navigli-2021-rebel-relation,lu-etal-2022-unified} and event extraction~\cite{li-etal-2021-document, lu-etal-2021-text2event} as generation tasks.
\citet{Nayak_Ng_2020} compared two models for the relation extraction task: copy mechanism-based decoding and text-to-text generation.
However, the results did not conclusively determine which method is superior.

\begin{table}[t]
    \centering
    \scriptsize
    \begin{tabular}{l|rrr}
    \toprule
         & \textbf{AAEC} &  \textbf{AbstRCT} & \textbf{CDCP}  \\
    \midrule
     \# component  & 6,089 & 3,279 & 4,931 \\
     \# relation & 3,832 & 2,060 & 1,220 \\
     \# components with multiple parents  & 0 & 31 & 160 \\
     \% words in nonargumentative span\tablefootnote{We used SpaCy 3.6.1 as a tokenizer for word counts.}  & 28.09 & 49.30 & 0\\
     \bottomrule
    \end{tabular}
    \caption{Statistics of AAEC, AbstRCT, CDCP.}
    \label{table:statistics_dataset}
\end{table}

\begin{table*}[t]
\footnotesize
\renewcommand{\arraystretch}{1.5} 
\begin{tabularx}{\textwidth}{lX}
\toprule
 Input Text & Advantages and disadvantages of the prevalent of English With the development of globalization , English became the dominated language in national trade , conference and many important events . This phenomenon has aroused a heated discussion in public . Some people claim that the prevalent of English brings a great number of benefits for people .  \\
\midrule
w/ nonargumentative span & Advantages and disadvantages of the prevalent of English With the development of globalization , English became the dominated language in national trade , conference and many important events . This phenomenon has aroused a heated discussion in public . Some people claim that [ the prevalent of English brings a great number of benefits for people \textbar \space claim for ] .  \\
w/o nonargumentative span & [ the prevalent of English brings a great number of benefits for people \textbar \space claim for ]  \\
\bottomrule
\end{tabularx}
  \caption{Example of input text and output in TANL and our format. The table shows that our output format reduces the number of tokens compared to the TANL format by removing tokens that do not contain any components or relations.}
  \label{table:eliminate_span_output_format}
\end{table*}

\begin{table*}[t]
\centering
  \includegraphics[width=\linewidth]{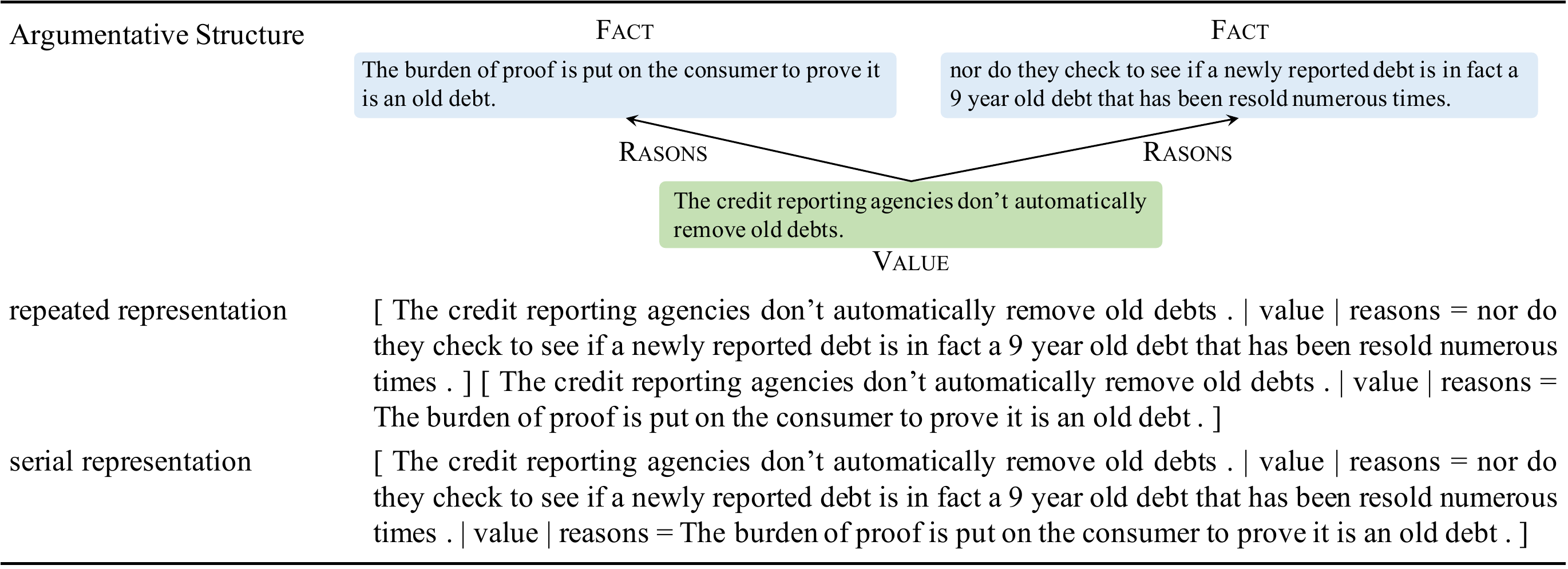}
  \caption{Examples of repeated representation and serial representation in CDCP.}
  \label{table:cdcp_output_format}
\end{table*}

Translation between Augmented Natural Languages~(TANL)~\cite{tanl} extends \citet{Nayak_Ng_2020}'s approach to text-to-text generation.
This methodology has proved highly effective for tasks such as relation extraction, named entity recognition, semantic role labeling, and coreference resolution.
This success can be attributed to the implementation of a more powerful pretrained encoder-decoder model, T5. 
In a recent study, \citet{10224326} proposed using T5 for sentence-level RST parsing as a form of text-to-text generation, demonstrating its effectiveness in analyzing sentence structures.
Their research motivates us to view document structure analysis, specifically argument mining, as more of a text generation task than a conventional natural language understanding task.
To address this text-to-text generation task, we propose using T5 within the TANL framework, which could be a significant solution.

\section{Proposed Methods}
Figure~\ref{fig:method_overview} provides an overview of our approach based on TANL \cite{tanl}.
To obtain the argumentatively annotated text for TANL, we align the original text with the given argumentative spans, their types, and relations.
Then, we fine-tune T5 with the TANL framework using the annotated texts.

\subsection{Task Formalization}
Given an input text $x$ consisting of $n$ words, it can be represented as $x=[x_1,\dots, x_n]$.
The objective of span identification is to extract spans $s =[x_{\mathrm{start}},\dots,x_{\mathrm{end}}]$ that include the argument.
Here, $\mathrm{start}$ and $\mathrm{end}$ indicate the indices marking the beginning and end of the span, respectively. 
Such extracted spans are denoted as $\mathrm{(start, end)}$.
Component classification labels the identified spans with component labels $c$ from a set $C$.
Set $C$ contains all component labels present in the dataset. 
As a result of the classification, components are represented as $(\mathrm{start},\mathrm{end},c)$.
Relation classification involves selecting source and target spans from the extracted spans and assigning a relation label $r$ from a predefined set $R$.
Set $R$ holds all relation labels in the dataset. The source and target spans are expressed as $(\mathrm{start^{src}}, \mathrm{end^{src}})$ and $(\mathrm{start^{tgt}}, \mathrm{end^{tgt}})$ respectively.
Consequently, the output from relation classification can be illustrated as $\mathrm{(start^{src}, end^{src}, start^{tgt}, end^{tgt}, r)}$.

\subsection{Argumentatively Annotated Text}
We adapt the output format of TANL's joint entity and relation extraction task to AM.
When a text span $s^{\mathrm{src}}$ with a specific component label $c$ depends on another text span $s^{\mathrm{tgt}}$ by a relation label $r$, we represent it as ``$[\;s^{\mathrm{src}} \; | \; c \; | \; r = s^{\mathrm{tgt}} \;]$''.
However, if the span $s^{\mathrm{src}}$ does not depend on others, we omit $s^{\mathrm{tgt}}$ and the relation label $r$, denoting it as ``$[ \; s^{\mathrm{src}} \; | \; c \; ]$''.
Below is an example illustrating the application of TANL's method to AM:

   \begin{quote}
   \textbf{\texttt{Input:}} For this reason , many marine lives have been endangered , in the extremes part of the reef become uninhabitable for these marine species . Thus , it is apparent that tourism has threatened the nature environments .
   \end{quote}
   \begin{quote}
   \textbf{\texttt{Output:}} For this reason ,  \textbf{[} \textit{many marine lives have been endangered , in the extremes part of the reef become uninhabitable for these marine species} | \textbf{premise} | \textbf{support} = \textcolor{black}{ \textit{tourism has threatened the nature environments} } \textbf{]} . Thus , it is apparent that \textbf{[} \textit{ tourism has threatened the nature environments} | \textbf{claim for} \textbf{]} .
   \end{quote}

\subsection{Elimination of Unnecessary Text Spans}
TANL attempts to annotate textual structure while maintaining integrity of  the original input text. 
However, our focus moves to tasks that require documents as input, in contrast to the original TANL's requirement for sentence-level inputs.
Reducing the maximum number of tokens in the encoder and decoder models is crucial for efficient computing.
As a result, we exclude nonargumentative spans from TANL’s annotation scheme.
Table~\ref{table:eliminate_span_output_format} shows examples of annotation with nonargumentative spans and without nonargumentative spans.

\subsection{Representation of Components with Multiple Parents}
The AM dataset contains components that depend on multiple parents. 
The output format of TANL's joint entity and relation extraction task cannot represent such structures in text, as it only adds annotations to the input text without repetition or deletion.
To address components with multiple parents, we employ two representations: \textbf{repeated representation} and \textbf{serial representation}\footnote{Serial representation is based on TANL’s nested entities and multiple relations output format}.
These examples are illustrated in Table~\ref{table:cdcp_output_format}.
As an illustration, a component with \textsc{Value} labeled ``\textit{The credit reporting agencies don't automatically remove old debts.}'' depends on two components labeled \textsc{Reasons}.
The repeated representation treats these as two separate relations and represents them sequentially, while the serial representation represents the first relation followed by the second.

\section{Experiments}

\subsection{Dataset}
We used three major benchmark datasets: Argument-annotated Essay Corpus~(AAEC)~\cite{stab-gurevych-2017-parsing}~, AbstRCT~\cite{mayer:hal-02879293} and the Cornell eRulemaking Corpus~(CDCP)~\cite{park-cardie-2018-corpus}.

\noindent\textbf{AAEC} includes annotations of components and relations for essays written by students.
It contains two types of data: essay-level and paragraph-level.
At the essay level, AM is performed on the entire essay as input, whereas at the paragraph level, AM is performed on the predefined paragraphs.
AAEC provides three component labels $C=\{\textsc{MajorClaim}$, $\textsc{Claim}$, $\textsc{Premise}\}$ and four relation labels $R=\{\textsc{For}, \textsc{Against}, \textsc{Support}, \textsc{Attack}\}$.
According to the AAEC annotation guidelines, a $\textsc{Claim}$ is always dependent on a $\textsc{MajorClaim}$.
Therefore, in our experiments, we adapted the labels to include four component labels $C=\{\textsc{MajorClaim}$, $\textsc{ClaimFor}$, $\textsc{ClaimAgainst}$, $\textsc{Premise}\}$ and two relation labels $R=\{ \textsc{Support}, \textsc{Attack}\}$.
During evaluation, we treat $\textsc{ClaimFor}$ and $\textsc{ClaimAgainst}$ as equivalent to $\textsc{Claim}$ and evaluate them as in the previous studies.
As shown in Table~\ref{table:statistics_dataset}, the ratio of words in the nonargumentative span to the total words is 28.09\%.
It is worth noting that a component does not have multiple parents.

\noindent\textbf{AbstRCT} is obtained from Randomized Controlled Trials~(RCT) from the MEDLINE for various diseases, such as neoplasm, glaucoma, hepatitis, diabetes, and hypertension.
It features three component labels $C=\{\textsc{MajorClaim}$, $\textsc{Claim}$, $\textsc{Evidence}\}$ and three relation labels $R=\{\textsc{Support}$, $\textsc{Attack}$, $\textsc{Partial-Attack}\}$.
As indicated in Table~\ref{table:statistics_dataset}, this dataset contains a significant proportion of nonargumentative spans.
Words in the nonargumentative spans account for 49.30\% of the total word count.

\noindent\textbf{CDCP} is annotated with components and relations for comments from citizens. 
It provides five component labels $C=\{\textsc{Fact}$, $\textsc{Testimony}$,
$\textsc{Value}$, $\textsc{Policy}$, 
$\textsc{Reference}\}$ and two relation labels $R=\{\textsc{Reasons}$, $\textsc{Evidence}\}$.
As shown in Table~\ref{table:statistics_dataset}, the CDCP does not contain any nonargumentative spans.
Furthermore, we observed that CDCP contains a greater number of components that depend on multiple other components than the other two datasets.

\subsection{Fine-tuning T5 with QLoRA}
In previous studies, TANL performed fine-tuning on the T5-Base by updating all parameters, a process known as \textit{full fine-tuning}.
However, in this study, we aim to explore the effects of additional parameters. 
To this end, we employ QLoRA~\cite{dettmers2023qlora} tuning to reduce GPU memory during the training of large parameter models such as T5-XL~(3B) and T5-XXL~(11B).
QLoRA is an adaptor that quantizes the model and applies Low-Rank Adapters~(LoRA)~\cite{hu2022lora} to it, helping reduce the number of parameters to be trained while maintaining performance levels comparable to full fine-tuning.

\begin{table*}[t]
\centering
\footnotesize
{\tabcolsep = 1.2mm
\begin{tabular}{lccccc}
\toprule
 & & \multicolumn{2}{c}{\textbf{Essay}} & \multicolumn{2}{c}{\textbf{Paragraph}} \\
\cmidrule(lr){3-4} \cmidrule(lr){5-6}
& \textbf{Params} & {\textbf{Component}} & {\textbf{Relation}} & {\textbf{Component}} & {\textbf{Relation}} \\
\midrule
ILP~\cite{stab-gurevych-2017-parsing} & {-} & {-} & {-}  & 62.61 & 34.74 \\
BLCC~\cite{eger-etal-2017-neural} & {-} & 63.23 & 34.82 & 66.69 & 39.83 \\
LSTM-ER~\cite{eger-etal-2017-neural} & {-} & 66.21 & 29.56 & 70.83 & 45.52 \\
BiPAM-syn~\cite{ye-teufel-2021-end} & 110M & {-} & {-} & {73.5 } & {46.4 } \\
BART-CPM~\cite{bao-etal-2022-generative} & {139M} & {-} & {-} & 75.94 & 50.08 \\
ST Model~\cite{morio-etal-2022-end} & 149M & 76.55 & 54.66 & 76.48 & 59.55 \\
\midrule
T5-Base  & 220M & 73.75 & 49.69 & 74.85 & 57.16 \\
T5-Large & 770M & 75.65 & 51.17 & 75.55 & 57.47 \\
T5-3B & 3B & 77.95 & 55.95 & 77.43 & 59.53 \\
T5-11B & 11B & 79.48 & 57.06 & 77.17 &59.02\\
\midrule
FLAN T5-Base & 220M & 75.17 &51.99 & 75.55 &58.51 \\
FLAN T5-Large & 770M &  77.75 &56.06 & 76.93 &58.57 \\
FLAN T5-XL & 3B & 78.51 &56.80 & 77.89 &60.94 \\
FLAN T5-XXL & 11B & \textbf{80.15} & \textbf{61.19} & \textbf{78.40} & \textbf{61.87} \\
\bottomrule
\end{tabular}
}
  \caption{Evaluation results at both the essay and paragraph levels obtained from AAEC. 
  ``Params'' indicates the model parameters of the pretrained language model used by each comparison model. \textbf{Bold} indicates the highest F1 score for each task.}
  \label{table:main_results}
\end{table*}

\begin{table}[t]
\centering
\footnotesize
\begin{tabular}{lcc}
\toprule
             &  \textbf{C} & \textbf{R}  \\
\midrule
ST Model~\cite{morio-etal-2022-end}       &   64.16   & 38.38 \\
\midrule
FLAN T5-Base      &  68.76  &  38.31 \\
FLAN T5-Large    &   71.11     &  44.47 \\
FLAN T5-XL    &    71.27    &  45.80 \\
FLAN T5-XXL      &   \textbf{72.86}     &  \textbf{47.66}\\
\bottomrule
\end{tabular}
  \caption{Evaluation results for Component-F1~(C) and Relation-F1~(R) in AbstRCT. \textbf{Bold} denotes the highest F1 score for each task.}
  \label{table:results_abstrct}
\end{table}

\begin{table}[t]
\centering
\footnotesize
\begin{tabular}{lcc}
\toprule
             &  \textbf{C} & \textbf{R}  \\
\midrule
BART-CPM~\cite{bao-etal-2022-generative} & 57.72  &  16.57\\
ST Model~\cite{morio-etal-2022-end}       &  68.90    & 31.94 \\
\midrule
FLAN T5-Base      &  66.80  & 23.19 \\
FLAN T5-Large    &    68.94    & 28.42 \\
FLAN T5-XL    &    72.12    &  31.01 \\
FLAN T5-XXL      &    \textbf{72.68}   & \textbf{33.96} \\
\bottomrule
\end{tabular}
  \caption{Evaluation results for Component-F1~(C) and Relation-F1~(R) in CDCP. \textbf{Bold} denotes the highest F1 score for each task.}
  \label{table:results_cdcp}
\end{table}

\subsection{Settings}
For the AAEC, we follow the train/dev/test split suggested by \citet{eger-etal-2017-neural}.
The number of essays for each dataset is 286, 36, and 80, while the number of paragraphs is 1587, 199, and 449, respectively.
For the AbstRCT, we used the neoplasm test set and adopted the split provided in the original paper~\cite{park-cardie-2018-corpus}, using 300 for training, 50 for dev, and 100 for testing. 
Following~\cite{niculae-etal-2017-argument}, we preserved 150 comments as a test set from the 731 comments within the CDCP.
For the CDCP, 15\% of training data were extracted as a dev set.

We examined T5~\cite{2020t5} and FLAN-T5~\cite{wei2022finetuned} as pretrained encoder-decoder models used within the TANL framework.
Our experiments were conducted with four different parameter models: Base~(220M), Large~(770M), XL~(3B), and XXL~(11B).
In each experimental setting, we report the average scores from three runs using different seeds.

\subsection{Compared Models}
We compared our method with the following models, including the state-of-the-art model:

\begin{itemize}
\item \textbf{ILP}~\cite{stab-gurevych-2017-parsing}: A feature-based method that employs Integer Linear Programming~(ILP) to parse each subtask in a pipelined fashion.
\item \textbf{BLCC}~\cite{eger-etal-2017-neural}: A method that treats Argument Mining (AM) as a sequence tagging problem.
\item \textbf{LSTM-ER}~\cite{eger-etal-2017-neural}: An end-to-end relation extraction model that uses LSTM-ER~\cite{miwa-bansal-2016-end}, which combines tree structure with sequential LSTM models.
\item \textbf{BiPAM-syn}~\cite{huang-etal-2021-document}: A model that employs BERT as a language model for end-to-end dependency parsing, incorporating biaffine operations and syntactic information.
\item \textbf{BART-CPM}~\cite{bao-etal-2022-generative}: An encoder-decoder model similar to ours, employing BART~\cite{lewis-etal-2020-bart} as the language model and the Constrained Pointer Mechanism (CPM).
\item \textbf{Single Task~(ST) model}~\cite{morio-etal-2022-end}: A state-of-the-art model in AM, employing a biaffine neural approach akin to BiPAM-syn, yet using Longformer~\cite{Beltagy2020Longformer} as the language model.
\end{itemize}

\begin{table*}[t]
\centering
\footnotesize
{\tabcolsep = 1.2mm
\begin{tabular}{llcccccc}
\toprule
 & & \multicolumn{2}{c}{\textbf{AAEC~(Essay)}} & \multicolumn{2}{c}{\textbf{AAEC~(Paragraph)}}  & \multicolumn{2}{c}{\textbf{AbstRCT}} \\
\cmidrule(lr){3-4} \cmidrule(lr){5-6} \cmidrule(lr){7-8}
\textbf{Output Format} & \textbf{Model} & {\textbf{Component}} & {\textbf{Relation}} & {\textbf{Component}} & {\textbf{Relation}} & {\textbf{Component}} & {\textbf{Relation}} \\
\midrule
\multirow{4}{*}{w/ nonargumentative span} &FLAN-T5 Base & 74.99 & 50.87 &74.97 & 57.54 & 65.30 &34.55 \\
                                  &FLAN T5-Large  & 77.76 & 55.62 & 76.53 & 59.09  & 69.47 & 39.66\\
                                  &FLAN T5-XL    &  78.73 & 57.21 & 77.17 & 61.03  & 73.13 &  42.39\\
                                  & FLAN T5-XXL  &  \textbf{80.59} & 60.37 & \textbf{79.06} & \textbf{62.38}  &  72.78 &   47.11\\
\midrule
\multirow{4}{*}{w/o nonargumentative span} &FLAN-T5 Base  & 75.17 &51.99 & 75.55 &58.51  & 68.76 &  38.31\\
                                  &FLAN T5-Large  & 77.75 &56.06 & 76.93 &58.57  & 71.11 &  41.49\\
                                  &FLAN T5-XL     & 78.51 &56.80 & 77.89 &60.94 & 71.27 &  45.80\\
                                  & FLAN T5-XXL   & 80.15 & \textbf{61.19} &  78.40 &61.87 &  \textbf{72.86} &  \textbf{47.66}\\
\bottomrule
\end{tabular}
}
  \caption{Comparison of Component-F1 and Relation-F1 scores with and without nonargumentative span output for essay-level and paragraph-level tasks in AAEC and AbstRCT.}
  \label{table:result_argument_non_argument}
\end{table*}

\begin{table*}[t]
\centering
\footnotesize
{\tabcolsep = 1.2mm
\begin{tabular}{llcccc}
\toprule
 & & \multicolumn{2}{c}{\textbf{Full Dataset}} & \multicolumn{2}{c}{\textbf{Multiple Parents}} \\
\cmidrule(lr){3-4} \cmidrule(lr){5-6}
\textbf{Output Format} & \textbf{Model} & {\textbf{Component}} & {\textbf{Relation}} & {\textbf{Component}} & {\textbf{Relation}} \\
\midrule
\multirow{4}{*}{repeated representation} &FLAN-T5 Base &  66.94  & 22.40 & 60.12 & 22.64 \\
                                  &FLAN T5-Large  & 66.80  & 23.19 & 64.15 & 27.43 \\
                                  &FLAN T5-XL    &  72.12    &  31.01 & 68.67 & 32.81 \\
                                  & FLAN T5-XXL  &  \textbf{72.68}   & 33.96 & \textbf{69.97} & 35.26 \\
\midrule
\multirow{4}{*}{serial representation} &FLAN-T5 Base  &   67.11  & 23.64 & 63.17 & 23.81 \\
                                  &FLAN T5-Large  & 67.57    & 30.36 & 65.18 & 33.93 \\
                                  &FLAN T5-XL     & 70.86    &  32.98 & 66.95 & 33.66 \\
                                  & FLAN T5-XXL   & 71.34   & \textbf{34.96} &  68.53 & \textbf{40.14} \\
\bottomrule
\end{tabular}
}
  \caption{Comparison of Component-F1~(C) and Relation-F1~(R) with different representations in CDCP. \textbf{Full dataset} shows results using all CDCP data, while \textbf{Multiple Parent} shows results using only data with multiple parents.}
  \label{table:result_cdcp_output_format}
\end{table*}

\subsection{Evaluation Measures}
Our methods were evaluated using the Component-F1 score and the Relation-F1 score\footnote{For the AAEC evaluation, we used the scripts of \citet{eger-etal-2017-neural}.}, which are considered the de-facto standard evaluation metrics in the field~\cite{stab-gurevych-2017-parsing,eger-etal-2017-neural,ye-teufel-2021-end, morio-etal-2022-end,bao-etal-2022-generative}.
Unlike AAEC, many studies have evaluated the component classification and the relation classification tasks of AbstRCT and CDCP given an oracle span~\cite{kuribayashi-etal-2019-empirical, morio-etal-2020-towards, mayer:hal-02879293}.
For the evaluation of AbstRCT and CDCP, we benchmarked our scores against those from prior research that also measured Component-F1 and Relation-F1 scores in an end-to-end fashion.

\subsection{Implementation Details}
We trained the model using a batch size of 32 for AAEC at the paragraph level, and a batch size of 8 for AAEC at the essay level, as well as for AbstRCT and CDCP.
We set the maximum token length to 512, for AAEC at the paragraph level, and 1,024 for the other datasets.
The learning rate for both the Base and Large models was set at 0.0005, while a learning rate of 0.0002 was used for the XL~(3B) and XXL~(11B) models.
All training took place on a single A100~(80GB) GPU over 10,000 steps, with checkpoints every 200 steps.

Typically, encoder-decoder models may not accurately replicate the input text, resulting in identified text spans that differ from those in the original text. 
To mitigate this issue, TANL uses the Needleman-Wunsch alignment algorithm~\cite{NEEDLEMAN1970443} to establish alignment between the output and input text spans. 
This process determines the position of words in the input text within the output text, an approach that we also adopted.

Following the method of \citet{bao-etal-2022-generative}, we performed inference using the development set with the fine-tuned model, selecting the best checkpoint based on the average scores of Component-F1 and Relation-F1. 
For training with QLoRA, we applied 4-bit quantization to the model and set the training hyperparameters as $r=16$ and $\alpha=32$. 
In addition, we trained the model by integrating the adapter into all linear layers.
Further details of our implementation can be found in Appendix~\ref{sec:appendix:implementation}.

\begin{figure}[t]
\centering
  \includegraphics[width=\linewidth]{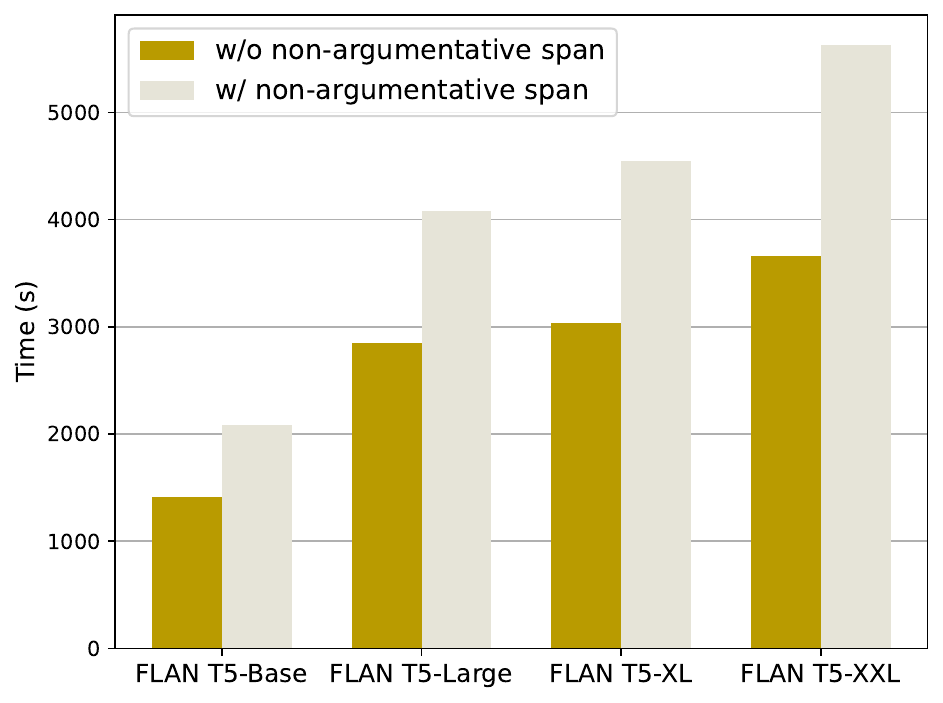}
  \caption{Comparison of inference time with and without nonargumentative spans in AbstRCT.
  We set the batch size to 2 for all models during inference and measured the time required to complete the process on the entire test dataset.
  }
  \label{fig:inference_time}
\end{figure}

\section{Results and Discussion}
\subsection{Main Results}
Table~\ref{table:main_results} shows the results obtained from the AAEC.
The results suggest that an increase in parameters tends to yield significant performance gains.
Although our models do not outperform state-of-the-art models when employing base or Large models, their performance is notably enhanced using models with billion-scale parameters.
Our model with 3B parameters (T5-XL and FLAN-T5-XL) achieved F1 scores comparable to current state-of-the-art models, while those with 11B parameters exceeded the existing top F1 scores.
Specifically, our FLAN T5-XXL model yielded Component-F1 scores of 80.15 at the essay level and 78.40 at the paragraph level, and Relation-F1 scores of 61.19 and 61.87, respectively.

Despite the larger parameter count, our straightforward model architecture using QLoRA proves practical.
The table also reveals that FLAN T5 outperforms T5, indicating that instruction-tuning on various tasks using FLAN~\cite{wei2022finetuned} has a positive impact on the AM task.

We also present the results obtained from AbstRCT and CDCP in Table~\ref{table:results_abstrct} and Table~\ref{table:results_cdcp}, respectively.
For AbstRCT, our FLAN T5-XXL achieved state-of-the-art performance with a Component-F1 score of 72.86 and Relation-F1 score of 47.66.
Similarly, the CDCP results also demonstrate state-of-the-art performance, with Component-F1 and Relation-F1 scores of 72.68 and 33.96, respectively. 
Our model significantly outperforms existing models in both datasets, reinforcing the effectiveness of our approach: text-to-text generation with the TANL framework.

\begin{figure*}[t]
\centering
  \includegraphics[width=\linewidth]{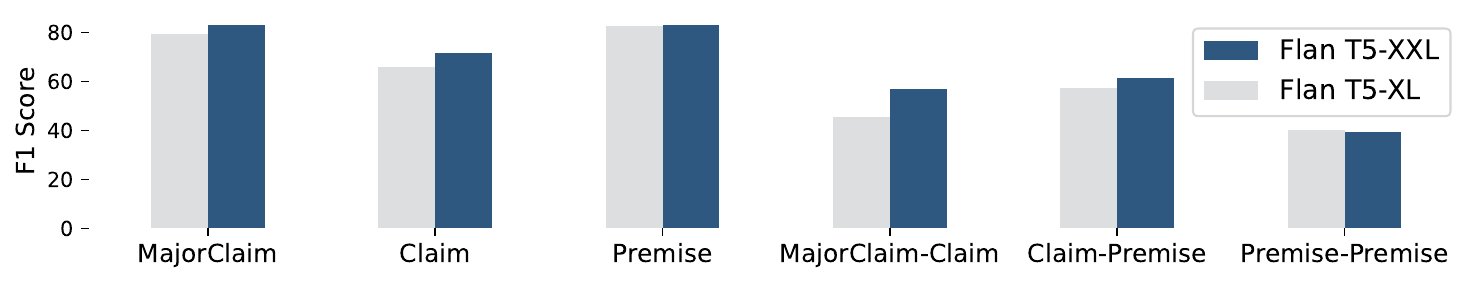}
  \caption{Comparison of F1 score output by label for the Component and Relation tasks. }
  \label{fig:labels_f1}
\end{figure*}

\begin{table*}[t]
\footnotesize
\renewcommand{\arraystretch}{1.5} 
\begin{tabularx}{\textwidth}{lX}
\toprule
Gold & [ the advertising expenses lead to a higher product price and some of them express fake information , creating information asymmetry between consumers and companies \textbar \space \textbf{claim against} ] [ its merits still outweigh these downsides \textbar \space premise \textbar \space attack = the advertising expenses lead to a higher product price and some of them express fake information , creating information asymmetry between consumers and companies ] \\
\midrule
    FLAN T5-XL &  [ the advertising expenses lead to a higher product price and some of them express fake information , creating information asymmetry between consumers and companies \textbar \space \textbf{\textcolor{red}{premise}} \textbar \space \textbf{\textcolor{red}{support = advertisements have no downsides}} ] [ its merits still outweigh these downsides \textbar \space premise \textbar \space attack = the advertising expenses lead to a higher product price and some of them express fake information , creating information asymmetry between consumers and companies ]  \\
\midrule
FLAN T5-XXL&  [ the advertising expenses lead to a higher product price and some of them express fake information , creating information asymmetry between consumers and companies | \textbf{claim against} ] [ its merits still outweigh these downsides | premise | attack = the advertising expenses lead to a higher product price and some of them express fake information , creating information asymmetry between consumers and companies ] \\
\bottomrule
\end{tabularx}
  \caption{Example of output text from FLAN T5-XL and FLAN T5-XXL.}
  \label{table:example_model_output}
\end{table*}

\subsection{Difficulties of Hyperparameter Tuning}
Our method offers an advantage over previous methods owing to the simplicity of our model.
Previous methods have individual hyperparameters for subtasks, including span identification, component classification, and relation classification.
\citet{morio-etal-2022-end} employed Optuna~\cite{Optuna} to determine the optimal hyperparameters.
However, hyperparameter tuning can become challenging owing to the interdependence of the three subtasks.
This complexity becomes more pronounced when using large models, potentially complicating the training process.

By contrast, as our method relies solely on the learning rate as its hyperparameter, it avoids the difficulties associated with hyperparameter tuning that are inherent in previous methods.
This simplicity is particularly advantageous when employing large-scale language models.

\subsection{Eliminating Nonargumentative Spans}
Our model does not output irrelevant text spans that cannot be classified as arguments.
To assess the impact of this elimination, we conducted evaluations of the model without eliminating the following three tasks\footnote{
We excluded CDCP from this evaluation as it does not contain irrelevant text spans.
}: AAEC at the essay level, AAEC at the paragraph level, and AbstRCT.

Table~\ref{table:result_argument_non_argument} shows the results.
Across all three tasks, it is evident that the performance is not degraded by the eliminations.
Furthermore, inference time can be reduced by eliminating irrelevant spans, an effect that is particularly pronounced in AbstRCT owing to its high count of nonargumentative spans.
Figure~\ref{fig:inference_time} illustrates the run time of the inference on AbstRCT.
The figure clearly shows the effectiveness of the eliminations.
Our method reduces the inference time by approximately 30\% for all models: Base, Large, XL, and XXL.

\subsection{Comparison of Component Representations with Multiple Parents}

Table~\ref{table:result_cdcp_output_format} presents a comparison of two different representations, repeated and serial, across the complete test dataset from the CDCP~(Full Dataset). 
The table also includes the Component-F1 and Relation-F1 scores, calculated only for components with multiple parents~(Multiple Parents).
The data demonstrates that there is no significant difference in performance between the two representations in the component classification.
However, in relation classification, the serial representation consistently outperforms the other models across all models. 
The results suggest that the repeated representation may not be optimal for tasks requiring the extraction of long-term relationships, such as relation classification.
This could be attributed to the model’s difficulty in capturing the full scope of relationships in the text, as the repeated representation breaks down one-to-many relationships into several separate one-to-one relations.

\subsection{Performance Improvement with Increasing Model Parameters}
In the AAEC task at the essay-level task, FLAN T5-XXL showed a significant improvement in Relation-F1 performance compared to FLAN T5-XL~(Table~\ref{table:main_results}).
The gain is surprisingly around 4.4 points.
To investigate the results in more detail, we discuss the F1 scores for each Component and Relation label.
Figure~\ref{fig:labels_f1} shows the results. 
For the component label, we observed a significant improvement in \textsc{Claim} (66.12 vs. 71.57) compared to \textsc{MajorClaim}  (79.34 vs. 82.96) and \textsc{Premise} (82.87 vs. 83.01).
When focusing on relation labels, we found the largest improvement is seen in the \textsc{MajorClaim}-\textsc{Claim} relation (45.33 vs.57.14).
According to the AAEC annotation rules, \textsc{Premise} to \textsc{Claim} and \textsc{Premise} to \textsc{Premise} are classified as relations within a paragraph, while \textsc{Claim} to \textsc{MajorClaim} can be a relation spanning different paragraphs.
Therefore, these results imply that FLAN T5-XXL can capture a longer dependency between arguments.
The substantial contribution of a large number of parameters to the improvement of distant dependency detection has a significant impact on the argument mining research community.

Table ~\ref{table:example_model_output} shows example outputs for gold, FLAN T5-XL, and FLAN T5-XXL.
In the table, FLAN T5-XL incorrectly predicts ``\textit{the advertising expenses lead to a higher product price and some of them express fake information , creating information asymmetry between consumers and companies}'' to \textsc{Premise} and depend on ``\textit{advertisements has no downsides}''.
On the other hand, FLAN T5-XXL correctly predicts that it is a \textsc{Claim} with \textsc{Against} relation to \textsc{MajorClaim}.

\section{Conclusion}

In this paper, we introduced a simple yet strong approach for argument mining~(AM) using the TANL framework for text-to-text generation.
To simplify and streamline annotations, we eliminated irrelevant text spans from the reference texts.
Experimental results obtained from  AAEC, AbstRCT, and CDCP demonstrated that our approach outperformed the current state-of-the-art method on these datasets.
Our research also indicated the efficacy of employing the TANL framework to predict document structures at the document level.
Furthermore, we found that removing irrelevant text spans decreased the inference time by approximately 30\% on AbstRCT.

\section*{Limitations}
Although our method achieves state-of-the-art Component-F1 and Relation-F1 scores across multiple datasets, its inference time poses a significant hurdle for practical implementation.
Inference time strongly relies on the length of the input text.
Although eliminating irrelevant spans can help reduce this time, our approach still necessitates longer inference time compared to previous methods.

Even though QLoRA reduces memory requirements during training, these large parameter models still require GPUs with a substantial memory capacity, such as the A100~(80GB).

Finally, we note that we only experimented with the TANL framework on encoder-decoder models such as T5 and FLAN T5.
Further research is necessary to verify whether our proposed method is compatible decoder-based Large Language models~(LLMs) such as GPT-4~\cite{openai2023gpt4} and LLAMA2~\cite{touvron2023llama}\footnote{As a preliminary experiment, we performed few-shot learning using GPT-4 turbo, however, the results were not satisfactory. Further details are provided in appendix~\ref{sec:appendix:decoder}.}.


\bibliography{anthology,custom}
\bibliographystyle{acl_natbib}

\appendix


\section{Implementation Details}
\label{sec:appendix:implementation}

\begin{table}[h]
\centering
\small
\begin{tabular}{lcc}
\toprule
         & \textbf{AAEC~(Paragraph)} & \textbf{AAEC~(Essay)} \\
\midrule
Batch size      &  32   & 8    \\ 
Max length      &  512  & 1024 \\
Step            & \multicolumn{2}{c}{10,000} \\
Dropout         & \multicolumn{2}{c}{0.1} \\
Adam beta1      & \multicolumn{2}{c}{0.9} \\
Adam beta2      & \multicolumn{2}{c}{0.998} \\
\bottomrule
\end{tabular}
\caption{Hyperparameters for AAEC.}
\label{table:model_parameter_aaec}
\end{table}

\begin{table}[h]
\centering
\small
\begin{tabular}{lcc}
\toprule
        & \textbf{AbstRCT} & \textbf{CDCP} \\
\midrule
Batch size      &  8   & 8    \\ 
Max length      & 1024 & 1024 \\
Step            & \multicolumn{2}{c}{10,000} \\
Dropout         & \multicolumn{2}{c}{0.1} \\
Adam beta1      & \multicolumn{2}{c}{0.9} \\
Adam beta2      & \multicolumn{2}{c}{0.998} \\
\bottomrule
\end{tabular}
\caption{Hyperparameters for AbstRCT and CDCP.}
\label{table:model_parameter_abstrct_cdcp}
\end{table}

\begin{table}[h]
\centering
\small
\begin{tabular}{lc}
\toprule
r      &   16  \\
lora alpha       &  32   \\
lora dropout     &   0.05   \\
bias      &    none   \\
task type & SEQ\_2\_SEQ\_LM \\
target modules   &  q, v, k, o, \\
&  wo, wi\_0, wi\_1  \\
load\_in\_4bit & True \\ 
bnb\_4bit\_quant\_type & nf4 \\
bnb\_4bit\_use\_double\_quant & True \\
bnb\_4bit\_compute\_dtype & torch.bfloat16 \\
\bottomrule
\end{tabular}
  \caption{Hyperparameters for fine-tuning with QeLoRA.}
  \label{table:qlora_parameter}
\end{table}

Our code implementation is based on TANL\footnote{\url{https://github.com/amazon-science/tanl}} and QLoRA\footnote{\url{https://github.com/artidoro/qlora}} scripts.
We used the T5\footnote{\url{https://huggingface.co/google-t5}} and FLAN T5\footnote{\url{https://huggingface.co/google/flan-t5-base}} models available on Hugging Face.

Table~\ref{table:model_parameter_aaec} and Table~\ref{table:model_parameter_abstrct_cdcp} detail the hyperparameters for fine-tuning.
We tuned the hyperparameters using the AAEC at the essay level.
The learning rates for Base, Large, XL~(3B), and XXL~(11B) models in both T5 and FLAN T5 were set at 0.0005, 0.0005, 0.0002, and 0.0002, respectively. 
These were determined through experimenting with all models, adjusting in increments of 0.0001 from 0.0001 to 0.0005.
It was observed that models with larger parameter sizes, such as XL (3B) and XXL (11B), showed improved performance with the lower learning rates. 
However, we confirmed that it was also possible to conduct training using the default learning rate of 0.0005, as suggested in the original TANL paper. 
Across all tasks, the Adam optimizer was employed. 
For QLoRA fine-tuning, we applied the same LoRA hyperparameters for all tasks, shown in Table~\ref{table:qlora_parameter}.

\section{Decoder-based Large Language Models}
\label{sec:appendix:decoder}

\begin{table}[h]
\centering
\footnotesize
\begin{tabular}{lcc}
\toprule
            &  \textbf{Component} & \textbf{Rrelation}  \\
\midrule
FLAN T5-Base      & 75.17   & 51.99 \\
FLAN T5-Large    &   77.75   & 56.06 \\
FLAN T5-XL    &      78.51  &  56.80 \\
FLAN T5-XXL      &    \textbf{80.15}   & \textbf{61.19} \\
\midrule
GPT4-turbo 3-shot      &    48.99   & 24.78\\
GPT4-turbo 5-shot      &    51.37  &  26.43 \\
GPT4-turbo 10-shot      &   52.79   & 26.76 \\
GPT4-turbo 20-shot      &   55.51   & 28.38 \\

\bottomrule
\end{tabular}
  \caption{Evaluation results of the few-shot learning using GPT-4-turbo. \textbf{Bold} denotes the highest F1 score for each task.}
  \label{table:decoder-llm-result}
\end{table}

Table~\ref{table:decoder-llm-result} shows  the outcomes from experiments using 3-shot, 5-shot, 10-shot, and 20-shot learning with GPT-4-turbo\footnote{Specifically, we used the  \texttt{gpt-4-1106-preview} version of the OpenAI API.}.
The Component-F1 and Relation-F1 scores for 3-shot learning were 48.99 and 24.78, respectively, while 20-shot learning improved to 55.51 and 28.38, respectively.
These scores are notably lower than those achieved with the fine-tuned FLAN T5 model. 
Although fine-tuning large language models (LLMs) is a promising direction for future research, finding the most effective prompts for such fine-tuning is still a challenge.
Consequently, our study did not extensively investigate the application of LLMs in Argument Mining (AM). 
Nonetheless, we believe that the text-to-text framework could be effectively integrated with decoders for AM tasks.

\end{document}